\documentclass{article}
\usepackage{spconf}
\usepackage{amsmath,graphicx}

\usepackage{multirow}
\usepackage{color}
\usepackage[dvipsnames]{xcolor}
\usepackage[normalem]{ulem}
\usepackage{booktabs}
\usepackage[super]{nth}
\usepackage[inline,shortlabels]{enumitem}
\usepackage{siunitx}
\usepackage{svg}
\usepackage{float}
\usepackage{amssymb}
\usepackage{url}
\usepackage{amsfonts,amssymb}
\usepackage[caption=false]{subfig}
\usepackage[linesnumbered,ruled,vlined]{algorithm2e}
 \usepackage{hyperref}

\usepackage{comment}

\DeclareMathOperator*{\argmin}{\arg\!\min}

\usepackage{todonotes}
\title{Speaker Diarization of Scripted Audiovisual Content}
\name{Yogesh Virkar, Brian Thompson, Rohit Paturi, Sundararajan Srinivasan, Marcello Federico}
\address{AWS AI Labs\\
{\small \tt yvvirkar@amazon.com}
}

\begin{document}
\ninept

\maketitle

\begin{abstract}

The media localization industry usually requires a verbatim script of the final film or TV production in order to create subtitles or dubbing scripts in a foreign language. In particular, the verbatim script (i.e. as-broadcast script) must be structured into a sequence of dialogue lines each including time codes,  speaker name and transcript. Current speech recognition technology alleviates the transcription step. However, state-of-the-art speaker diarization models still fall short on TV shows for two main reasons: (i) their inability to track a large number of speakers, (ii) their low accuracy in detecting frequent speaker changes. To mitigate this problem, we present a novel approach to leverage production scripts used during the shooting process, to extract pseudo-labeled data for the speaker diarization task. We propose a novel semi-supervised approach and demonstrate improvements of 51.7\% relative to two unsupervised baseline models on our metrics on a 66 show test set.

\end{abstract}


\section{Introduction \& Related Work}
\label{sec:intro}
Media localization is the process of adapting audiovisual content across languages and cultures to reach international audiences. Media localization is a very labor-intensive process \cite{chaume_audiovisual_2014, brannon2022dubbing}, where complexity and cost depend, among other factors, mainly on the chosen localization modality (subtitling, voiceover or dubbing) and the required quality.  Among the initial localization steps is the creation of a so called as-broadcast-script, which mainly consist in a verbatim transcript of the audio, structured into dialogue lines in paragraph form, each annotated with character (speaker)\footnote{To conform with naming conventions in the speech community, we will henceforth refer to speaker and character interchangeably, although there is a clear distinction between the two concepts. } name and timing information. In the media and entertainment industry, a significant amount of localization deals with scripted content, such as movies, television series, and documentaries.  Recent progress in speech technology has contributed in reducing the labor costs of the transcription process by providing drafts that can be post-edited much faster than transcribing from scratch. 
However, there is much room to improve on segmenting and labeling the transcript with speaker and timing information. This process falls under the scope of speaker diarization technology, which addresses the question of "who spoke when" inside a given audio file or stream.

Despite the significant progress on speaker diarization using end-to-end neural diarization models \cite{9003959, 9746225, edaeend_20_interspeech, 9747301}, clustering-based approaches based on speaker embeddings are still the most popular for handling long audios with more than 4 speakers \cite{bwedaeend_21_icassp}. However, as we show, conventional clustering-based techniques using even the state-of-the-art  speaker embeddings like ECAPA-TDNN \cite{ecapatdnn_21_interspeech} and Resnet \cite{he_2016_cvpr} architectures, still fall short of delivering useful speaker diarization performance for media localization. This for two main reasons: the high number of speakers that need to be tracked inside a movie or TV show and the required precision in detecting speaker changes.  

\begin{figure}[t]
  \centering
  \includegraphics[width=0.9\linewidth]{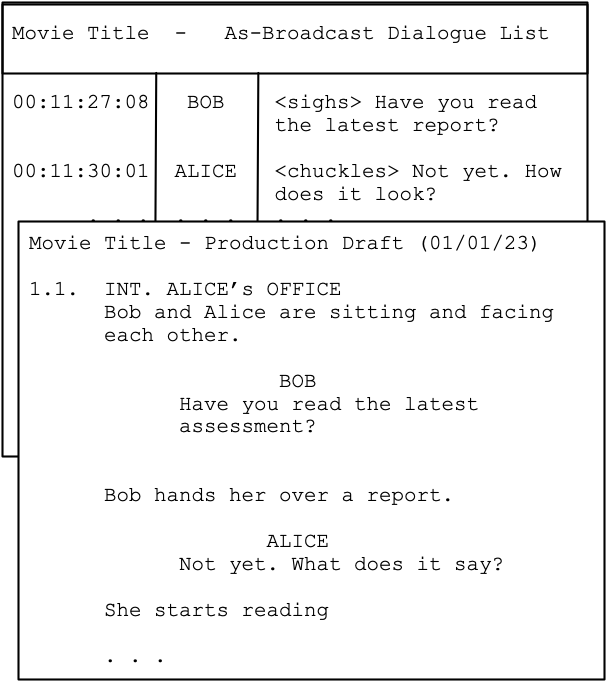}
  \caption{Illustrative example showing the difference between as-broadcast and production scripts.}
  \label{fig:scripts}
\end{figure}

In this work, we investigate methods for improving speaker diarization of audiovisual content for which we assume to have a production script available. Briefly, a production script is a version of the screenplay used during the production of the show. It guides the director and actors while performing or shooting, but can be subject to many changes during the production: dialogue lines can be deleted, changed or moved to a different position. The closest literature similar to this is the target-speaker voice activity detection (TS-VAD) \cite{ivan_20_interspeech}, where we are given (or we infer) voice samples for each speaker, and we would like to detect whenever these speakers are present in the audio. However, this work is different in that the speaker labels form the production scripts are noisy and may not cover all speakers in the show, and we want to leverage the power of speaker embeddings instead of training a TS-VAD from scratch. 

In the context of creating an as-broadcast script for the  post-production of a movie or a TV show, we frame speaker diarization as the process of obtaining character names with corresponding timing information for each dialogue line. 
Inputs of this process \cite{case2013film} include the final cut video, the audio stem file, a clean version of the final audio mix with no background music or sound effects, and the production script. As-broadcast scripts are verbatim transcripts of the final edited version of the production and are often used to produce subtitles, and translations for dubbing. The as-broadcast script can be seen as a revised and enriched version of the production script. \autoref{fig:scripts} shows an example of a production script and corresponding as-broadcast script. As shown, the as-broadcast script also includes time codes and annotations of sound effects and on screen graphics. The core part, in common with the production script, is the arrangement of the transcript into dialogue lines with the corresponding speaker. However, on our data, we empirically measured on average an exact match of dialogue lines between as-broadcast and production scripts to be only around 10\%.

Constrained spectral clustering has been used in the context of speaker diarization \cite{wang_speaker_2018, xia2022turn},
but these constraints relate to speaker \emph{turns} - that is, pairs of frames which can and cannot come from the same speaker. In contrast, in our work we locate sub-segments where we can determine the speaker a-prior with high confidence and our constraint is the actual speaker label. 

Prior works have also considered using a two-step clustering method of TV shows \cite{7078606} to first cluster speakers within each scene and then combine speakers from multiple scenes. While it can help to perform multi-modal diarization using lexical \cite{park19_interspeech, paturi2023lexical}, visual \cite{6380624, sharma2022using, bredin_gregory_2016, otsuka2008realtime, noulas2011multimodal, vallet2012multimodal, liu22t_interspeech, xavier_av_sd_2015} or longitudinal \cite{villalba2015variational, ferras2016system} information, we leave this for future work.

First, we automatically extract overlapping dialogue lines between production script and ASR transcript and use them as pseudo labels to inform speaker diarization. 
Then, we introduce a novel semi-supervised version of the spectral clustering method. 
We report results on a proprietary test set of 66 English TV shows. Our approach improves on our metrics on average by +XX\% over a strong unsupervised baseline.

\section{Method}

\subsection{Extraction of Pseudo-labeled Data}
\label{subsec:pldata}

To utilize the production scripts, we detect sections of the script that match the final audio with high confidence. 
To do this, we first perform ASR using an in-house system to extract a noisy transcript with word-level timestamps. Note that we do not use segmentation from our ASR model as do not want to bias our pseudo-labels with potentially erroneous segmentation coming from ASR. 
Instead, we perform alignment between dialogue lines in the production scripts to \emph{words} in the ASR transcript using the
Vecalign\footnote{\url{https://github.com/thompsonb/vecalign}} \cite{thompson-koehn-2019-vecalign, thompson-koehn-2020-exploiting}
sentence alignment toolkit.

We search for alignments of sizes \{1,1\}, \{1,2\}, \dots \{1,50\}; that is, we allow each dialogue line to align with up to 50 words in the ASR transcript.
We allow for deletions of both dialogue lines and words from the transcript, to account for dialogue lines that are missing/added to the production script, respectively. For Vecalign, we empirically set a deletion percentile fraction of 0.015 in order to find good-quality alignments. 

For each dialogue line that is aligned to one or more words in the ASR transcript, we take the start time of the first ASR word and the end time of the last ASR word to get a time range where the character associated with the given dialogue line is speaking. These character names and time ranges are used as pseudo labels in the following sections.

\subsection{Speaker Diarization}\label{sec:speakerdiarization}
 Similar to prior works on speaker diarization\footnote{We refer the reader to \cite{park2022review} for a comprehensive and up to date survey on this topic.} \cite{wang_speaker_2018, xia2022turn, zhang2019fully}, our speaker diarization models are based on first extracting embeddings for speech segments. We first run an in-house Voice Activity Detector (VAD) to obtain speech segments from audio. These are further subdivided into uniform sub-segments of 1s duration. Each sub-segment is transformed into 512 dimensional embeddings using a speaker embedding model. The speaker embedding model follows a ResNet34 architecture \cite{he_2016_cvpr} and is trained with a combination of classification and metric loss \cite{33chung2020defence} with 12k speakers and 5k hours of data.

\subsubsection{Unsupervised Speaker Diarization Method (Baseline)}
\label{subsec:unsupervisedmodel}

Let $X = [x_1, x_2, \dots, x_n]$ denote the speaker embeddings for $n$ sub-segments extracted from the given input audio. These embeddings are used to construct an affinity matrix, $A$, such that $A_{ij}$ denotes the cosine similarity between the embeddings $x_i$ and $x_j$. We perform a series of refinements on the Affinity matrix to both smooth and denoise the data followed by Spectral Clustering as outlined in \cite{wang_speaker_2018}. The number of speaker clusters, $k=\widetilde{k}$ where $\widetilde{k}$ is automatically determined by the maximum eigen-gap of the eigenvalues in the Spectral Clustering step. 

\subsubsection{Semi-supervised Speaker Diarization Method}
\label{subsec:semisupervisedmodel}

Following the spectral clustering step of the unsupervised model, let $v_1, v_2, \dots, v_{k}$ be the eigenvectors corresponding to the $k$ largest eigenvalues. For the $i^{th}$ sub-segment, we obtain the corresponding spectral embedding $e_i = [v_{1i}, v_{2i}, \dots, v_{ki}]$. In order to cluster the embeddings, we replace the unsupervised K-means algorithm and propose a semi-supervised version that can utilize the prior information as outlined in Algorithm~\ref{alg:constrained_kmeans}.

We consider as inputs the spectral embeddings $[e_1, e_2, \dots, e_n]$ and pseudo labels $[l^{'}_1, l^{'}_2, \dots, l^{'}_n]$ for each of n input audio sub-segments. If $i^{th}$ sub-segment does not have a pseudo label, we assign $l^{'}_i=0$. Additional inputs are $\widetilde{k}$, i.e., the estimated number of speakers using eigen-gap method and $k^{'}$ that denotes the number of unique pseudo labels or known speakers. Since $\widetilde{k}$ is often underestimated (see \autoref{subsec:pldata100}), in step 1, we set number of speakers $k$ as the maximum of $\widetilde{k}$ and $k^{'}$. For the constrained K-means algorithm, we first initialize the cluster centroids using prior information from pseudo labels in lines 2-4. Let the cluster centroids be denoted as $[\mu_1, \mu_ 2, \dots, \mu_k]$. 
For all the $k^{'}$ known speakers we compute the centroids by averaging the corresponding embeddings. In steps 5-6, the remaining $k - k^{'}$ speaker centroids using the standard K-means++ algorithm \cite{david_sergei_2007_kmeans_plusplus}. For the E-step of the K-means we compute the label assignments in lines 8-12. For pseudo-labeled sub-segments we do not change label assignments and label only the unknown sub-segments. The idea is to not change the label assignment of known speakers in order to bias the clustering algorithm. Finally, lines 13-14 denote the M-step of K-means in order to recompute the speaker centroids. We repeat the E and M steps until convergence. 

\newcommand\mycommfont[1]{\footnotesize\ttfamily\textcolor{black}{#1}}
\SetCommentSty{mycommfont}
\SetNoFillComment
\SetKwComment{Comment}{/* }{ */}
\RestyleAlgo{ruled}
\begin{algorithm}
\caption{Constrained K-means}\label{alg:constrained_kmeans} 
\KwData{$[e_1, e_2, \dots, e_{n}]$, $[l^{'}_1, l^{'}_2, \dots, l^{'}_n]$, $\widetilde{k}$, $k^{'}$}
\KwResult{$[l_1, l_2, \dots, l_{n}]$}

\tcc{Initialization}
Set $k=\max\left(\widetilde{k}, k^{'}\right)$

Initialize $k$ centroids $[\mu_1, \mu_2, \dots, \mu_k]$

\For{$j \gets 1$ \KwTo $k^{'}$}{
    $\mu_j = \displaystyle\frac{\sum_{i=1}^{n} e_i . \mathrm{1}_{l^{'}_i=j}}{\sum_{i=1}^{n}\mathrm{1}_{l^{'}_i=j}}$
}
\For{$j \gets k^{'} + 1$ \KwTo $k$} {
    Initialize $\mu_j$ using the standard K-means++
}

\tcc{Clustering}
\Repeat{convergence}{
    \tcc{E-Step}
    \For{$i \gets 1$ \KwTo $n$} {
        \If{$l^{'}_i \neq 0$} {
            $l_i = l^{'}_i$
        }
        \Else {
            $l_i = \argmin_j \sum_{j=1}^{k} || e_i - \mu_j ||^2$ 
        }
    }
    \tcc{M-Step}
    \For{$j \gets 1$ \KwTo $k$} {
        $\mu_j = \displaystyle\frac{\sum_{i=1}^{n} e_i . \mathrm{1}_{l_i=j}}{\sum_{i=1}^{n}\mathrm{1}_{l_i=j}}$
    }
}
\end{algorithm}

\section{Evaluation Data \& Metrics}
\label{sec:evaldatametrics}

For evaluation, we collected a test set of 66 episodes from 21 shows from a major studio such that for each episode we have a production script, an audio speech stem and the ground-truth as-broadcast script. This is a total of 36.3 hours of episode runtime with 880 distinct speakers and 36429 total dialogues lines with shows spanning diverse genres such as drama, comedy, suspense, and kids. Note that the distribution of speech time across the 880 speakers is highly skewed making it a challenging test set for this task.

In order to obtain the ground-truth data for speaker diarization, we first segment the audio stem at the dialogue level using the timing information available in the as-broadcast script. For each dialogue segment, we additionally run an in-house voice activity detector in order to correctly identify speech segments. All speech segments corresponding to the same dialogue are annotated with the same speaker label.

We use the following automatic metrics for evaluating speaker diarization: 
\begin{enumerate}
    \item Diarization Error Rate (DER) is the standard metric for comparing speaker diarization systems and consists of three components: false alarm, missed detection and speaker error.
    \item Speaker Change Detection F1 (SCD) that defines the F1 score for correctly identifying the time boundary between speaker turns under some tolerance \cite{hang2022_icassp}. This is particularly important since not identifying the correct speaker changes can result in more time-consuming and expensive post-editing process in order to obtain quality as-broadcast scripts. 
\end{enumerate}

\section{Experiments \& Results}
\label{sec:experiments}

\begin{table}[t]
\caption{Model performance on diarization error rate (DER) and speaker change detection F1 (SCD). Significance testing is done at level $p<0.01$ against Unsupervised ($*$) and Pyannote ($+$) models.}
\label{tab:autoeval}
\centering
\begin{tabular}{@{}lccc@{}}
\toprule
Model                            & DER $\downarrow$      & SCD $\uparrow$ \\\midrule
Unsupervised                     & 48.99                 & 32.53          \\ 
Unsupervised ($k=k^{'}$)         & 53.55                 & 38.14          \\ 
Pyannote                         & 40.65                 & 34.36          \\
Semi-supervised [this work]    & \enspace{  }\textbf{27.04}$^{*+}$  & \enspace{  }\textbf{54.86}$^{*+}$      \\\bottomrule
\end{tabular}
\end{table}

In this section, we conduct multiple experiments to evaluate the performance of our models described in \autoref{sec:speakerdiarization}. Due to a small test set of 66 episodes, we do not perform tuning using our in-domain test set. Instead, for all the models, the hyperparameters for spectral clustering namely thresholding factor and thresholding percentile were tuned to minimize DER on the validation sets of Dihard \cite{dihard_dev} and ICSI \cite{icsi}. Our unsupervised model achieves an DER of 3.5\% on the test split of AMI \cite{carletta2006ami}. As an additional baseline, we compare against the publicly-available Pyannote speaker diarization \cite{Bredin2021} pipeline\footnote{\url{https://huggingface.co/pyannote/speaker-diarization}}.

We use the unsupervised model described in \autoref{subsec:unsupervisedmodel} and Pyannote speaker diarization pipeline as two baseline systems. To make the results comparable to these unsupervised baseline models, for the semi-supervised model we convert the predicted speaker names to speaker labels and report results using the Hungarian assignment \cite{kuhn1955hungarian} for all models.

First, we test the performance of pseudo-labeled data extraction process as described in \autoref{subsec:pldata}. In particular, we find that on average, with high confidence we are able to label 10.9\% dialogue lines over the entire test set. Further, by comparing with the ground-truth as-broadcast script, the pseudo labels are found to be 74.5\% accurate.

Next, we conduct two sets of experiments. In the first set, we compare the baseline approaches with the semi-supervised model (\autoref{subsec:semisupervisedmodel}) utilizing the entire available pseudo-labeled data. In the second set of experiments, we vary the amount of psuedo-labeled data in order to assess the impact on the proposed semi-supervised method. We report the Diarization Error Rate (DER) and Speaker Change Detection F1 (SCD). For SCD, unlike a more relaxed tolerance of 200ms as considered in \cite{hang2022_icassp}, we use a more conservative value of 100ms since correcting speaker changes is quite time-consuming and expensive.

\subsection{Speaker Diarization Results}

\begin{figure}[t]
  \centering
  \includegraphics[width=\linewidth]{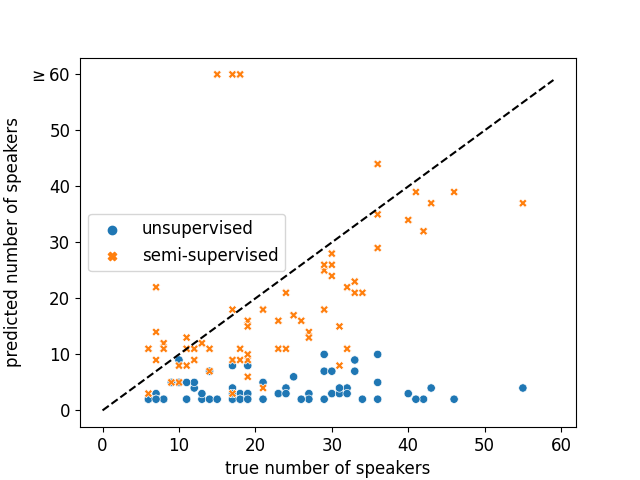}
  \caption{True vs predicted number of speakers for unsupervised and semi-supervised models with x=y shown as dashed black line.}
  \label{fig:error_est_k}
\end{figure}
\label{subsec:pldata100}

As shown in \autoref{tab:autoeval}, for our unsupervised baseline system obtains a poor DER of 48.99\% and a poor SCD of 32.53\%. This is primarily due to a large number of speakers typically found in TV shows and the lack of prior knowledge to handle them. Additionally, we find that the spectral clustering algorithm highly underestimates the number of clusters (or speakers), thereby worsening speaker confusion and consequently the DER. 
This is illustrated in \autoref{fig:error_est_k} that shows the scatter plot of true (x-axis) vs predicted (y-axis) number of speakers for unsupervised and semi-supervised models. The dashed line shows y=x line. As shown, the semi-supervised approach gets closer to the true number of speakers than the unsupervised approach. This underestimation also results in a low recall in identifying speaker changes thereby resulting in a poor SCD. 

The high number of predicted speakers, i.e., $\geq 60$, for the semi-supervised approach comes from production scripts overestimating $k^{'}$. On inspecting on a handful episodes, we find that these errors are mostly due to large changes in the script during the shooting process (see \autoref{sec:intro}) as well as noise in production scripts such as lack of character names being normalized.

Next, we correct the predicted number of speakers for the unsupervised approach by fixing $k=k^{'}$. However, as shown in \autoref{tab:autoeval}, while this improves SCD by +17.2\% relative it worsens DER by 9.3\% relative. Thus the lack of prior knowledge is a limiting factor for performance. 

The publicly available Pyannote speaker diarization pipeline based on Bayesian HMM clustering of x-vectors (VBx) \cite{Bredin2021} improves over our baseline approach on both DER and SCD by relative +17.0\% and +5.6\% respectively. 

On the other hand, using prior information, i.e., using the entire pseudo labeled data helps the semi-supervised method to outperform our unsupervised baseline with statistically significant\footnote{Significance testing is done at level $p<0.01$ using the two-sample Student's t-test.} relative improvements of +44.8\% and +68.6\% on DER and SCD respectively. Semi-supervised approach also improves over Pyannote pipeline by relative +33.5\% and +59.7\% on the same metrics. 

\subsection{Ablation Study}
\begin{figure}[]
  \centering
  \includegraphics[scale=0.4, width=\linewidth]{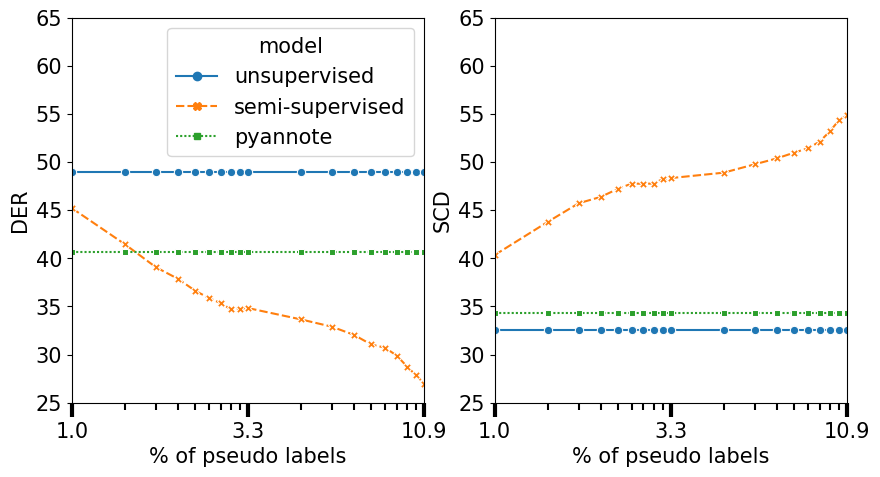}
  \caption{Performance of semi-supervised model by varying amount of available pseudo-labeled data on DER (left) and SCD (right).}
  \label{fig:varying_pldata}
\end{figure}

The amount of pseudo-labeled data depends heavily on how similar the final recording is to the production script. 
In \autoref{fig:varying_pldata}, we vary the amount of pseudo-labeled data and plot DER (left panel) and SCD (right panel) for the proposed semi-supervised method. This simulates what would happen if the shows were more dissimilar to the production script. Performance of our unsupervised model and the Pyannote diariaztion pipeline are shown for reference. 

We find that even using small amount of pseudo labels, helps the semi-supervised approach beat both the baseline models. Using 3\% psuedo labels improves performance over our unsupervised baseline and Pyannote pipeline respectively by relative 20.3\% and 3.9\%. For SCD, the improvements are stronger as using only 1\% of pseudo labels helps obtain relative improvements of 24.0\% and 17.4\%. Finally, adding more pseudo labels helps the model achieve consistent performance improvements for both metrics.

\section{Conclusions}
\label{sec:conclusions}
In this paper, we focus on the problem of speaker diarization for the media localization industry that requires a verbatim script of the the final film in order to localize content in particular foreign languages. While the current state-of-art speech recognition technology works reasonably well for transcription, it is unable to cope with large number of speakers for the problem of speaker diarization.  
We propose a novel approach to extract 
pseudo-labels from 
drafts of final scripts, also called as production scripts. We then present a novel semi-supervised speaker diarization method based on constrained clustering that is able to utilize these pseudo labels in order to vastly improve performance over a strong unsupervised baseline model and the publicly available Pyannote diarization pipeline. Our proposed approach shows strong performance improvements on average over both baselines and considered metrics by relative +51.7\%.

\bibliographystyle{IEEEbib-abbrev.bst}
\bibliography{biblio.bib}
\end{document}